\documentclass[runningheads]{llncs}
\usepackage{graphicx}
\usepackage{epstopdf}

\usepackage{amsmath,amssymb} 
\usepackage{color}
\usepackage{amssymb}
\usepackage{enumerate}
\usepackage[table,xcdraw]{xcolor}
\makeatletter
\newcommand*{\rom}[1]{\expandafter\@slowromancap\romannumeral #1@}
\makeatother
\newcommand{\etal}{\textit{et al}.}
\usepackage{hyperref}
\hypersetup{
  linkcolor  = red,
  citecolor  = green,
  urlcolor   = blue,
  colorlinks = true,
}
\begin{document}

\title{Learning Rotation Adaptive Correlation Filters in Robust Visual Object Tracking} 
\titlerunning{Rotation Adaptive Correlation Filters} 


\author{Litu Rout\inst{1} \and
	Priya Mariam Raju\inst{1} \and
	Deepak Mishra\inst{1} \and
	Rama Krishna Sai Subrahmanyam Gorthi\inst{2}}
%

\authorrunning{L. Rout et al.} 


\institute{Department of Avionics, Indian Institute of Space Science and Technology\\ Thiruvananthapuram, Kerala, India - 695 547 \\
	\email{liturout1997@gmail.com, priyamariyam123@gmail.com, deepak.mishra@iist.ac.in} \and
	Department of Electrical Engineering, Indian Institute of Technology\\
	Tirupati, Andhra Pradesh, India - 517 506\\
	\email{rkg@iittp.ac.in}}
\maketitle

\begin{abstract}
Visual object tracking is one of the major challenges in the field of computer vision. Correlation Filter (CF) trackers are one of the most widely used categories in tracking. Though numerous tracking algorithms based on CFs are available today, most of them fail to efficiently detect the object in an unconstrained environment with dynamically changing object appearance. In order to tackle such challenges, the existing strategies often rely on a particular set of algorithms.  Here, we propose a robust framework that offers the provision to incorporate illumination and rotation invariance in the standard Discriminative Correlation Filter (DCF) formulation. We also supervise the detection stage of DCF trackers by eliminating false positives in the convolution response map. Further, we demonstrate the impact of displacement consistency on CF trackers. The generality and efficiency of the proposed framework is illustrated by integrating our contributions into two state-of-the-art CF trackers: SRDCF and ECO. As per the comprehensive experiments on the VOT2016 dataset, our top trackers show substantial improvement of $14.7\%$ and $6.41\%$ in robustness, $11.4\%$ and $1.71\%$ in Average Expected Overlap (AEO) over the baseline SRDCF and ECO, respectively. \footnote{The final authenticated version is available online \href{https://doi.org/10.1007/978-3-030-20890-5_41}{here}.}

\end{abstract}
\section{Introduction}
\label{sec:intro}
Visual object tracking finds applications in diverse fields like traffic monitoring, surveillance systems, human computer interaction etc. Though the same object is being tracked throughout a given video sequence, the conditions under which the video is captured may vary due to changes in the environment, object, or camera. Illumination variations, object deformations, object rotations etc. are various challenges that occur due to changes in the aforementioned factors. A good tracking algorithm should continue tracking a desired object and its performance should remain unaffected under all these conditions.
Most of the existing trackers can be classified as either generative or discriminative. The generative trackers~\cite{bib1,bib2,bib3,bib4,bib5} use the object information alone to search for the most probable region in an image that matches the initially specified target object. On the other hand, the discriminative trackers~\cite{bib6,bib7,bib8,bib9,bib10,huang2017learning,fan2017parallel} use both the object and background information to learn a classifier that discriminates the object from its background. The discriminative trackers, to a large extent, make use of CFs as classifiers. The main advantage of CFs is that correlation can be efficiently performed in the Fourier domain as simple multiplication, as proven by Parseval's theorem. For this reason, CF trackers are learned and all computations are performed efficiently in the Fourier domain that leads to drastic reduction in computational complexity~\cite{bib8}. Thus, the CF trackers have gained popularity in the community because of their strong discriminative power, which emerges due to implicit inclusion of large number of negative samples in training.

Despite all the advancements in CF tracking, most of these algorithms are not robust enough to object deformations, rotations, and illumination changes. These limitations are due to the inherent scarcity of robust training features that can be derived from the preceding frames. This restricts the ability of the learned appearance model to adapt the changes in target object. Therefore, we propose rotation adaptiveness and illumination correction schemes in order to extract sophisticated features from previous frames that helps in learning robust appearance model. The rotation adaptiveness, up to some extent, tackles the issues of object deformation due to the robustness in representation.

The main contributions of this paper are as follows. \textbf{(a)} An Illumination Correction filter (IC) (Sec.~\ref{ic}) is introduced in the tracking framework that eliminates the adverse effects of variable illuminations on feature extraction.  \textbf{(b)} We propose an approach to incorporate rotation adaptiveness (Sec.~\ref{racf}) in standard DCF by optimizing across the orientations (Sec.~\ref{oo}) of the target object in the detector stage. The orientation optimization helps in extracting robust features from properly oriented bounding boxes unlike most state-of-the-art trackers that rely on axis aligned bounding boxes.  \textbf{(c)} Building on it, we supervise the sub-grid localization cost function (Sec.~\ref{fsgd}) in the detector stage of DCF trackers. This cost function is intended to eliminate the false positives during detection. \textbf{(d)} Further, we show the impact of enhancing smoothness through displacement correction (Sec.~\ref{dc}), and demonstrate all these contributions on two popular CF trackers: Spatially Regularized Disriminative Correlation Filters (SRDCF)~\cite{bib10}, and Efficient Convolution Operators (ECO)~\cite{bib18}. 
%
%
%
%

\begin{figure}[!t]
	\begin{center}
		\includegraphics[height=6cm,width=0.9\linewidth ]{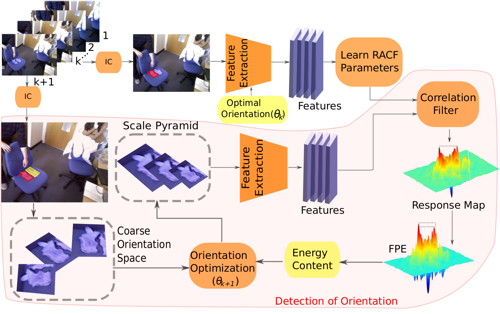}
	\end{center}
	\caption{As the pipeline indicates, both train ($k^{th} $) and test ($k+1^{th}$) frames undergo illumination correction (IC) prior to feature extraction. The training features are then used to learn the parameters of Rotation Adaptive Correlation Filter (RACF). During detection stage, each candidate patch passes through a coarse orientation space from which the orientation optimizer picks a seed orientation. The seed orientation is usually the object's immediate previous orientation which is then used by Newton's iterative optimization scheme as initial point to determine optimal orientation for $k+1^{th}$ frame. The optimizer maximizes the total energy content in the False Positive Eliminated (FPE) convolutional response map. The response map corresponds to the winning scale in the scale pyramid. Note that the optimal orientation in the first frame ($\theta_{1}$) is assumed to be $0^{\circ}$ without loss of generality. Thereafter, the optimal orientations in the subsequent frames are determined through a deterministic optimization strategy.}   
	\label{fig:fg1}
\end{figure}

Though we have demonstrated the importance of our contributions by integrating with SRDCF and ECO, the proposed framework is generic, and can be well integrated with other state-of-the-art correlation filter trackers. The rest of the paper is structured as following. At first we discuss the previous works related to ours (Sec.~\ref{related}), followed by the illumination correction filter (Sec.~\ref{ic}), the detailed description of rotation adaptive correlation filters (Sec.~\ref{racf}), and displacement consistency (Sec.~\ref{dc}). Thereafter, we provide experimental evidence (Sec.~\ref{exp}) to validate our contributions. In Fig.~\ref{fig:fg1}, we show the pipeline of our overall architecture.

\section{Related Works} \label{related}
Numerous variants of the CF tracker have been proposed by adding constraints to the basic  filter design, and by utilizing different feature representations of the target object. Initial extensions start with the KCF tracker~\cite{bib8} which uses a kernel trick to perform efficient computations in the Fourier domain. The Structural CF tracker~\cite{bib9} uses a part based technique in which each part of the object is independently tracked using separate CFs.  Danelljan \etal~\cite{bib10} proposed the SRDCF tracker which uses a spatial regularizer to weigh the CF coefficients in order to emphasize the target locations and suppress the background information. Thus, the SRDCF tracker includes a larger set of negative patches in training, leading to a much better discriminative model. 

The earlier trackers directly used the image intensities to represent the target object. Later on, feature representations such as color transformations\cite{bib11,bib12,bib13,bib8}, Colornames\cite{bib14} etc. were used in CF trackers. Due to significant advancement of deep neural networks in object detection, features from these networks have also found applications in tracking, giving rise to substantial gain in performance. The deep trackers, such as DeepSRDCF~\cite{bib15}, MDNet~\cite{bib21}, and TCNN~\cite{bib22}, clearly indicate the distinctive feature extraction ability of deep networks. The HCF tracker~\cite{bib16} exploits both semantic and fine-grained details learned from a pre-trained Convolutional Neural Network (CNN). It uses a multi-level correlation map to locate the target object. The CCOT tracker~\cite{bib17} uses DeepSRDCF~\cite{bib15} as the baseline and incorporates an interpolation technique to learn the filter in continuous domain with multi-resolution feature maps. The ECO tracker~\cite{bib18} reduces the computational cost of CCOT by using a factorized convolution operator that acts as a dimensionality reduction operator. ECO also updates the features and filters after a predefined number of frames, instead of updating after each frame. This eliminates redundancy and over-fitting to recently observed samples. As a result, the deep feature based ECO tracker does reasonably well on diverse datasets outperforming other CF trackers by a large margin.

Among rotation adaptive tracking, Zhang~\etal~propose an exhaustive template search in joint scale and spatial space to determine the target location, and learn a rotation template by transforming the training samples to Log-Polar domain, as explained in RAJSSC~\cite{zhang2015joint}. We learn rotation adaptive filter in the cartesian domain by incorporating orientation in the standard DCF, unlike exhaustive search. In contrast to a recent rotation adaptive scheme SiameseFC-DSR, as proposed by Rout \etal~\cite{rout2018rotation}, we incorporate rotation adaptiveness directly in the standard DCF formulation, by performing a pseudo optimization on a coarse grid in the orientation space, leading to robust training of CF. Qianyun \etal~\cite{du2015rotation} use a multi-oriented Circulant Structure with Kernel (CSK) tracker to get multiple translation models each dominating one orientation. Each translation model is built upon the KCF tracker. The model with highest response is picked to estimate the object's location. The main difference is that we do not learn multiple translation models at various orientations, as proposed in multi-oriented CSK, in order to reduce computational cost. In contrast, we optimize the total energy content in convolution responses at the detector stage with respect to object's orientation. The multi-channel correlation filter is then learned from a set of training samples which are properly oriented through a deterministic approach. Note that our training process requires a single model.


\section{Illumination Correction (IC) Filter } \label{ic}
Illumination changes occur in a video due to dynamically changing environmental conditions, such as waving tree branches, low contrast regions, shadows of other objects etc. This variable illumination gives rise to low frequency interference, which is one of the prominent causes of disturbing the object's appearance. As the appearance of an object changes dramatically under different lighting conditions, the learned model fails to detect the object, leading to reduction in accuracy and robustness. Also, we may sometimes be interested in high frequency variations, such as edges, which are part of the dominant features in representing an object. Though these issues are investigated extensively in image processing community, to our knowledge, necessary attention for the same is not paid explicitly, even in the state-of-the-art trackers. Though deep features have shown to be fairly
invariant to random fluctuations in input image, such as blur, white noise, illumination variation etc. the experimental results in Sec.~\ref{exp} shows that trackers with deep features also fail to  track the object under these challenges. Therefore, we intend to introduce Illumination Correction filter~(IC) in the tracking paradigm in order to tackle the aforementioned issues up to some degree without affecting the usual tracked scenarios. At first, we employ a standard contrast stretching mechanism~\cite{bib19} to adjust the intensities of each frame. The contrast stretched image is then subjected to unsharp masking~\cite{bib19}, a popular image enhancement technique in order to suppress the low frequency interference, and enhance high variations. To our surprise, the performance of the baseline trackers improves by a considerable amount just by enhancing the input images, as given in Sec.~\ref{exp}. This validates the fact that the robust feature extractors still lack high quality visual inputs, which otherwise can lead to substantial gain in performance. 


\section{Rotation Adaptive Correlation Filters (RACF)} \label{racf}
Here, we elaborate the training and detection phase of rotation adaptive correlation filters in light of standard SRDCF. For the ease of understanding and clearly distinguishing our contributions, we have used identical notations as in SRDCF~\cite{bib10}. First, we explain standard SRDCF training and detection process, and then, we integrate rotation adaptiveness with false positive elimination into the optimization framework of CF, unlike heuristic template search~\cite{zhang2015joint,rout2018rotation}.

\subsection{SRDCF Training and Detection} \label{std_srdcf}
In the standard DCF formulation, a multi-channel correlation filter $f$ is learned from a set of training samples $\left \{ \left ( x_{k},y_{k} \right ) \right \}_{k=1}^{t}$. Each training sample $x_{k}$ has a $d$-dimensional feature map, which is extracted from an image region. All the samples are assumed to be of identical spatial resolution $M \times N$. Thus, we have a $d$-dimensional feature vector $x_{k}(m,n) \in \mathbb{R}^{d}$ at each spatial location $(m,n) \in \Omega := \left \{0,\dots,M-1\right \} \times \left \{0,\dots,N-1\right \}$. We also denote feature layer $l \in \left \{1,\dots,d \right \}$ of $x_{k}$ by $x_{k}^{l}$. The target of each training sample $x_{k}$ is denoted as $y_{k}$, which is a scalar valued function over the domain $\Omega$. The correlation filter $f$ has a stack of $d$ layers, each of which is a $M \times N$ convolution filter $f^{l}$. The response of the convolution filter $f$ on a $M \times N$ sample $x$ is computed by, 
\begin{equation} \label{sfx}
S_{f}(x)=\sum_{l=1}^{d} x^{l} * f^{l}.
\end{equation}
Here, $*$ represents circular convolution. The desired filter $f$ is obtained by minimizing the $L^2$-error between convolution response $S_{f}(x_{k})$ of training sample $x_k$ and the corresponding label $y_k$ with a more general Tikhonov regularizer $w : \Omega \rightarrow \mathbb{R}$,
\begin{equation} \label{err1}
\varepsilon \left ( f \right ) = \sum_{k=1}^{t}\alpha_k\left \| S_f(x_k) - y_k \right \|^2 + \sum_{l=1}^{d} \left \| \frac{w}{MN} \cdot f^{l} \right \|^2.
\end{equation}
Here, $\cdot$ denotes point-wise multiplication. With the help of Parseval's theorem, the filter $f$ can be equivalently computed by minimizing the equation~(\ref{err1}) in the Fourier domain with respect to Discrete Forurier Transform (DFT) coefficients $\hat{f}$, 
\begin{equation} \label{err1f}
\hat{\varepsilon}(\hat{f}) = \sum_{k=1}^{t}\alpha_k\left \| \sum_{l=1}^{d}\hat{x}_k^l \cdot\hat{f}^l - \hat{y_k} \right \|^2 + \sum_{l=1}^{d}\left \| \frac{\hat{w}}{MN}*\hat{f}^l \right \|^2.
\end{equation}
Here, $ \hat{} $ denotes the DFT of a function. After learning the DFT coefficients $\hat{f}$ of filter $f$, it is typically applied in a sliding-window-like manner on all cyclic shifts of a test sample $z$. Let $\hat{s} :=\mathcal{F}\left \{ S_f(z) \right \} = \sum_{l=1}^{d}\hat{z}^l \cdot \hat{f}^l$ denote DFT ($\mathcal{F}$) of the convolution response $S_f(z)$ evaluated at test sample $z$. The convolution response $s(u,v)$ at continuous location $(u,v) \in [0,M) \times [0,N)$ are interpolated by, 
\begin{equation} \label{sub-grid1}
s(u,v)=\frac{1}{MN}\sum_{m=0}^{M-1}\sum_{n=0}^{N-1}\hat{s}\left ( m,n \right )e^{i2\pi \left ( \frac{m}{M}u + \frac{n}{N}v \right )}.
\end{equation}
Here, $i$ denotes the imaginary unit. The maximal sub-grid location $(u^*,v^*)$ is then computed by optimizing $\text{arg max}_{(u,v) \in [0,M) \times [0,N)}s\left ( u,v \right )$ using Newton's method, starting at maximal grid-level score $(u^{(0)},v^{(0)}) \in \Omega$. In a nutshell, the standard SRDCF adapts translation invariance efficiently by exploiting the periodic assumption with spatial regularization, but this does not learn rotation adaptiveness inherently. Therefore, we propose to extend the discriminative power of SRDCF by learning rotation adaptive filters through a deterministic optimization procedure.

\subsection{RACF Training and Detection}
First, we incorporate rotation adaptiveness in spatially regularized correlation filters by learning from appropriately oriented training samples. Similar to SRDCF, we solve the resulting optimization problem in the Fourier domain, by employing a deterministic orientation in each training sample. Let $\theta_k$ denotes the orientation corresponding to $x_k$. Without loss of generality, it can be assumed that $\theta_{k}=0, \forall k \leq 1$. The training sample $x_k$ undergoes rotation $\theta_k$ by,
\begin{equation} \label{xktheta}
x_k^{\theta}\left ( m,n \right )_{(m,n) \in \Omega} = \left\{\begin{matrix}
x_k\left ( {m}',{n}' \right )&,& \left ( {m}',{n}' \right ) \in \Omega  \\
0&,& elsewhere 
\end{matrix}\right.
\end{equation}
where $\left ( m,n \right )$ and $\left ( {m}',{n}' \right )$ are related by, 
\begin{equation}
\begin{bmatrix}
n\\ 
m
\end{bmatrix}=\begin{bmatrix}
\cos(\theta_k)  & -\sin(\theta_k)\\ 
\sin(\theta_k)& \cos(\theta_k)
\end{bmatrix}\begin{bmatrix}
{n}'\\
{m}' 

\end{bmatrix}.
\end{equation}
In other words, $x_k^{\theta}$ is obtained by rotating $x_k$ anti-clockwise with an angle $\theta_k$ in the Euclidean space and cropping same size $M \times N$ as $x_k$. In order to avoid wrong gradient estimation due to zero paddings, we use a common solution that bands the rotated image patch with cosine window. This does not disturb the structure of the object assuming that the patch size is larger than the target object. This is different from standard SRDCF, in a sense that we learn the multi-channel correlation filter $f$ from properly oriented training samples $\left \{ \left ( x_{k}^{\theta},y_{k} \right ) \right \}_{k=1}^{t}$. The training stage of rotation adaptive filters is explained in the following Sec.~\ref{train}.

\subsection{Training: Learning RACF parameters} \label{train}
The convolution response $S_f(x_k^{\theta})$ of the rotated training sample $x_k^{\theta} \in \mathbb{R}^{d}$ is computed by,
\begin{equation} \label{sfx2}
S_f(x_k^{\theta})=\sum_{l=1}^{d} x_k^{\theta l} * f^{l}.
\end{equation}

After incorporating rotation into the DCF formulation, the resulting cost function is expressed as, 
\begin{equation} \label{err2}
\varepsilon_{\theta} \left ( f \right ) = \sum_{k=1}^{t}\alpha_k\left \| S_f(x_k^{\theta}) - y_k \right \|^2 + \sum_{l=1}^{d} \left \| \frac{w}{MN} \cdot f^{l} \right \|^2.
\end{equation}

Similar to SRDCF, we perform the Gauss-Seidel iterative optimization in Fourier domain by computing DFT of equation~(\ref{err2}) as, 
\begin{equation} \label{err2f}
\hat{\varepsilon}_{\theta}(\hat{f}) = \sum_{k=1}^{t}\alpha_k\left \| \sum_{l=1}^{d}\hat{x}_k^{\theta l} \cdot\hat{f}^l - \hat{y_k} \right \|^2 + \sum_{l=1}^{d}\left \| \frac{\hat{w}}{MN}*\hat{f}^l \right \|^2.
\end{equation}
The equation~(\ref{err2f}) is vectorized and simplified further by using fully vectorized real-valued filter, as implemented in the standard SRDCF~\cite{bib10}. The aforementioned training procedure is feasible, provided we obtain the object's orientation corresponding to all the training samples beforehand. In the following Sec.~\ref{detect}, we propose an approach to localize the target object and detect its orientation by optimizing a newly formulated objective function.

\subsection{Detection: Localization of the Target Object} \label{detect}
At the detection stage, the correlation filter $f$ learned from $t$ training samples are utilized to compute the convolution response of a test sample $z$ obtained from $(t+1)^{th}$ frame, which is then optimized to locate the object in that $(t+1)^{th}$ frame. For example, at $t=1$, we learn the coefficients of $f$ from $(x_{k=1}^{\theta = 0^{\circ}},y_{k=1})$ and detect the object location, $(u^*_{k+1},v^*_{k+1})$, and orientation, $\theta_{k+1}$ in the $(t+1)^{th}$, i.e., $2^{nd}$ frame. For efficient detection of scale, we construct multiple resolution test samples $\left \{ z_r \right \}_{r \in \left \{ \left \lfloor \frac{1-S}{2} \right \rfloor,\dots,\left \lfloor \frac{S-1}{2} \right \rfloor \right \}}$ by resizing the image at various scales $a^{r}$, as implemented in SRDCF\cite{bib10}. Here, $S$ and $a$ denote the number of scales and scale increment factor, respectively. Next, we discuss the false positive elimination scheme, which offers notable gain in the overall performance.

\subsubsection{False Positive Elimination (FPE)} \label{fpe}
\begin{figure}[!t]
	\begin{center}
		\includegraphics[height=4.5cm,width=0.8\linewidth ]{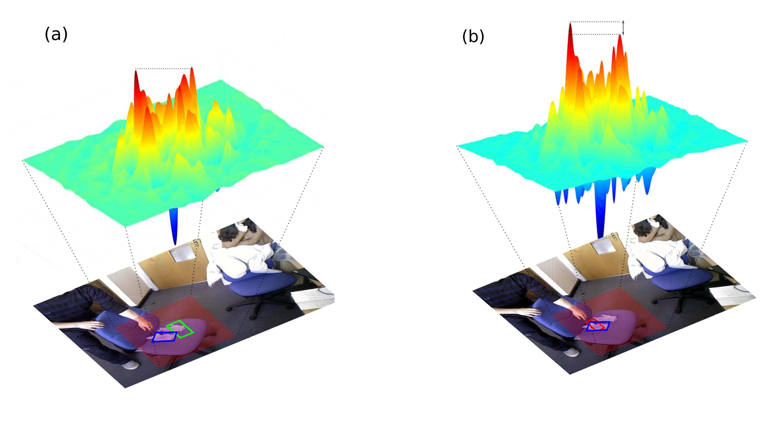}
	\end{center}
	\caption{Sample frames from the sequence glove of VOT2016~\cite{VOT_TPAMI}. The blue, green, and red rectangle shows the output of groundtruth, ECO, and F-ECO (with FPE), respectively. Convolution response of shaded (red) region obtained directly (a) without, and (b) with optimization through false positive elimination.}  
	\label{conv_resp}
\end{figure}

As per our extensive experiments, we report that the convolution response map of test sample may sometimes contain multiple peaks with equal detection scores. This situation usually arises when the test sample is constructed from an image region that consists of multiple objects with similar representations as target object. In fact, this issue can occur in many real world scenarios, such as glove, leaves, rabbit etc. sequences from VOT2016 dataset~\cite{VOT_TPAMI}. Therefore, we propose to maximize $\frac{s(u,v)}{\left \| \left ( u-u^*_k,v-v^*_k \right ) \right \|}$ unlike SRDCF, which focuses on maximizing $s(u,v)$ alone. Here, $(u^*_k,v^*_k)$ denote the sub-grid level target location in the $k^{th}$ frame. Thereby, we intend to detect the object that has high response score as well as minimum deviation from previous location. Arguably, this hypothesis is justified by the fact that it is less likely for an object to undergo drastic deviation from immediate past location. Due to identical representation in feature space, both the gloves have equal response score as shown in Fig.~\ref{conv_resp}(a). However, the FPE scheme mitigates this issue, as shown in Fig.~\ref{conv_resp}(b), by maximizing the response score subject to minimum deviation from previous centroid which in turn creates a very distinct decision boundary. Note that the Gauss-Seidel optimization of total energy content through FPE directly yields this scalar valued response (Fig.~\ref{conv_resp}(b)) without any post-processing.

\subsubsection{Detection of Orientation (DoO)} \label{oo}
Here, we elaborate the detection mechanism of object's orientation in the test sample. Let $\hat{s}_\theta :=\mathcal{F}\left \{ S_f(z^\theta) \right \} = \sum_{l=1}^{d}\hat{z}^{\theta l} \cdot \hat{f}^l$ represents the DFT ($\mathcal{F}\left \{. \right \}$) of convolution response $S_f(z^\theta)$, evaluated at $\theta$ orientation of test sample $z$. Similar to equation~(\ref{sub-grid1}), we compute $s_\theta(u,v)$ on a coarse grid $(u,v) \in \Omega$ by,
\begin{equation} \label{sub-grid2}
s_\theta(u,v)=\frac{1}{MN}\sum_{m=0}^{M-1}\sum_{n=0}^{N-1}\hat{s}_\theta\left ( m,n \right )e^{i2\pi \left ( \frac{m}{M}u + \frac{n}{N}v \right )}.
\end{equation}
Then, the aim is to find orientation that maximizes the total energy content in the convolution response map by,
\begin{equation} \label{thetak1}
\theta_{k+1} = \text{arg max}_{\theta \in \Phi  }\left \{ \sum_{u=0}^{M-1}\sum_{v=0}^{N-1}\left ( \frac{S_\theta\left ( u,v \right )}{\left \| \left ( u-u^*_k,v-v^*_k \right ) \right \|} \right )^2 \right \}.
\end{equation}
Here, $\Phi:=\left \{ \theta_k \pm a\delta  \right \}$, where $a = 0,1,2,\dots, A $. Thus, the orientation space $\Phi$ consists of $(2A+1)$ number of rotations with step size $\delta$. In our experiments, we have used $\delta = 5^{\circ}$, and $A = 2$ based on the fact that an object's orientation is less likely to change drastically between consecutive frames. Nevertheless, the orientation can be further optimized by Newton's approach, or any suitable optimization algorithm, starting at optimal coarse orientation $\theta_{k+1}$. Also, a suitable combination of $A$ and $\delta$ can be chosen for searching exhaustively in $\Phi$, but at the expense of time complexity. Next, we incorporate the FPE and DoO techniques in Fast Sub-grid Detection method of standard SRDCF (Sec.~\ref{fsgd}) formulation.

\subsubsection{Fast Sub-grid Detection} \label{fsgd}
We apply the Newton's optimization strategy, as in SRDCF, for finding the sub-grid location that maximizes the detection score. However, we incorporate the false positive elimination and optimal orientation in the standard SRDCF sub-grid detection. Thus, we compute the sub-grid location that corresponds to maximum detection score by,
\begin{equation} \label{fsgdeq}
\left ( u^*_{k+1},v^*_{k+1} \right ) = \text{arg max}_{\left ( u,v \right )\in [0,M) \times [0,N)}\left \{ \frac{S_{\theta_{k+1}}(u,v)}{\left \| \left ( u-u^*_k,v-v^*_k \right ) \right \|} \right \},
\end{equation}
starting at $(u^{(0)},v^{(0)}) \in \Omega$, such that $\left \{ \frac{S_{\theta_{k+1}}(u^{(0)},v^{(0)})}{\left \| \left ( u^{(0)}-u^*_k,v^{(0)}-v^*_k \right ) \right \|} \right \}$ is maximal.
\section{Displacement Consistency} \label{dc}
Motivated by the displacement consistency techniques, as proposed in~\cite{rout2018rotation}, we enhance the degree of smoothness imposed on the movement variables, such as speed and angular displacement. We update the sub-grid location, $\left ( u^*_{k+1},v^*_{k+1} \right )$ obtained from equation~(\ref{fsgdeq}) by,
\begin{equation} \label{mceq}
\begin{split}
\left ( u^*_{k+1},v^*_{k+1} \right ) &= \left ( u^*_{k},v^*_{k} \right ) + d_{1n}\angle \varphi_{1n},\\ d_{1n} &= \omega_{d} \times d_{1} + \left ( 1-\omega_d \right ) \times d_0,\\
\varphi_{1n} &= \omega_a \times \varphi_1 + \left ( 1-\omega_a \right )  \times \varphi_0,
\end{split}
\end{equation}
where, $d_0 = \left \| \left ( u^*_k - u^*_{k-1},v^*_k - v^*_{k-1} \right ) \right \|$, $d_1 = \left \| \left ( u^*_{k+1} - u^*_{k}, v^*_{k+1} - v^*_{k} \right ) \right \|$,\\ $\varphi_0 = \arctan \left ( u^*_k-u^*_{k-1},v^*_k-v^*_{k-1} \right )$, $\varphi_1 = \arctan \left ( u^*_{k+1}-u^*_{k},v^*_{k+1}-v^*_{k} \right ), \omega_d = 0.9, \omega_a=0.9$. The abrupt transition from $\left ( u^*_{k},v^*_{k} \right )$ to $\left ( u^*_{k+1},v^*_{k+1} \right )$ is restricted by reducing the contribution of $d_1$ and $\varphi_1$ slightly to $0.9$. Note that for $\omega_d=\varphi=1$, the updated $\left ( u^*_{k+1},v^*_{k+1} \right )$ of equation~(\ref{mceq}) remains unaltered from the optimal solution of equation~(\ref{fsgdeq}). In the following Sec.~\ref{exp}, we briefly describe our experimental setup, and critically analyze the results.
\section{Experiments} \label{exp}
First, we detail the experimental setup, and then carry out the ablation studies to analyze the effect of each individual component towards overall tracking performance. Then we conduct extensive experiments to compare with various state-of-the-art trackers both qualitatively and quantitatively on VOT2016~\cite{VOT_TPAMI} and OTB100~\cite{WuLimYang13} benchmark. In all our experiments, we use VOT toolkit and OTB toolkit for evaluation on VOT2016 and OTB100 benchmark, respectively. 
\subsection{Implementation Details}
In order to perform an unbiased analysis that may arise due to varying numerical precision of different systems, we evaluate all the models, including baseline SRDCF and ECO on the same system under identical experimental setup. All the experiments are conducted on a single machine: Intel(R) Xeon(R) CPU E3-1225 v2 @ 3.20GHz, 4 Core(s), 4 Logical Processor(s) and 16GB RAM. The proposed tracker has been implemented on MATLAB with Matconvnet. We use the similar parameter settings as baseline, apart from the additional parameters $\delta = 5^{\circ}$, and $A = 2$ in rotation adaptive filters. These settings are selected because the orientation does not change drastically between consecutive frames. In IC, we use output intensity range $[0,255]$ for contrast stretching, and a threshold $0.5$ for unsharp masking. This intensity range is selected so as to match with the conventional representation (uint8) of most images. On the other hand, this threshold is selected manually by observing qualitative results on VOT2016. 


\subsection{Estimation of Computational Complexity}
The Fast Fourier Transform (FFT) of a 2-dimensional signal of size $ M \times N $ can be computed in $\mathcal{O} (MN \log MN)$. Since there are $d$ feature layers, $S$ scales, and $(2A+1)$ orientations, the training and detection stage of our algorithm requires $\mathcal{O}\left ( ASdMN \log MN \right )$ FFT computations. To compute the convolution response, the computed FFTs require $\mathcal{O}\left ( ASdMN \right )$ multiplication operations, and $\mathcal{O}\left ( ASMN \right )$ division operations. The division operations are used in False Positive Elimination (FPE) strategy. Assuming that the Newton's optimization converges in $N_{Ne}$ iterations, the total time complexity of matrix multiplication and FPE sums up to $\mathcal{O}\left ( \left (ASdMN+ASMN  \right )N_{Ne} \right )$. In contrast to standard SRDCF~\cite{bib10}, we learn the multi-resolution filter coefficients from properly oriented training samples. After detection of orientation through optimization of total energy content on a coarse grid, the training samples are oriented appropriately in $\mathcal{O}\left ( MN \right )$ time complexity. The fraction of non-zero elements in $A_t$ of size $dMN \times dMN$, as given in standard SRDCF, is bounded by the upper limit $\frac{2d+k^2}{dMN}$. Thus, the total time complexity of standard SRDCF training, assuming that the Gauss-Seidel optimization coverges in $N_{GS}$ iterations, sums up to $\mathcal{O}\left ( \left ( d+k^2 \right )dMNN_{GS} \right )$. In addition to the standard SRDCF training, our approach requires $\mathcal{O}\left ( MN \right )$ operations to orient the samples, leading to a total complexity of $\mathcal{O}\left (MN+ \left ( d+k^2 \right )dMNN_{GS} \right )$. Therefore, the overall time complexity of our RIDF-SRDCF is given by,
\begin{equation}
\mathcal{O}\left (ASdMN\log MN + \left ( ASdMN + ASMN \right )N_{Ne}+ MN+ \left ( d+k^2 \right )dMNN_{GS} \right )
\end{equation}
and that of SRDCF is given by,
\begin{equation}
\mathcal{O}\left (dSMN\log MN + SMNN_{Ne} + \left ( d+k^2 \right )dMNN_{GS} \right )
\end{equation}
Note that the overall complexity of both of these are largely dominated by $\mathcal{O}\left (\left ( d+k^2 \right )dMNN_{GS} \right )$, leading to only slight increment in computational cost due to the additional terms, but resulting in significant improvement in overall performance of RIDF-SRDCF relative to standard SRDCF. In fact, the RIDF-SRDCF runs at 5 fps and SRDCF runs at 7fps on our machine.

\subsection{Ablation Studies}
We progressively integrate Displacement consistency (D), False positive elimination (F), Rotation adaptiveness (R), Illumination correction (I), and their combinations into ECO framework for faster experimentaion, and assimilate the impact of each individual component on AEO, which is the standard metric on VOT2016 benchmark. We evaluate each Ablative tracker on a set of 16 videos (Table~\ref{eco_comp}) during the development phase. The set is constructed from the pool of 60 videos from VOT2016 dataset. A video is selected if its frames are labelled as either severe deformation, rotation, or illumination change. Note that the FPE scheme improves the performance in every integration, and illumination correction alone provides a gain of 7.7\% over base RDF-ECO. As per the results in Table~\ref{eco_comp}, the proposed ideas independently and together provide a good improvement relative to base model.

\begin{table}[!h]
	\centering
	\caption{Quantitative evaluation of Ablative trackers on a set of 16 challenging videos from VOT2016 benchmark.}
	\label{eco_comp}
	\begin{tabular}{lllllllll}
		\hline
		Tracker & ECO      & \begin{tabular}[c]{@{}l@{}}D-\\ ECO\end{tabular} & \begin{tabular}[c]{@{}l@{}}DF-\\ ECO\end{tabular} & \begin{tabular}[c]{@{}l@{}}R-\\ ECO\end{tabular} & \begin{tabular}[c]{@{}l@{}}RF-\\ ECO\end{tabular} & \begin{tabular}[c]{@{}l@{}}RD-\\ ECO\end{tabular} & \begin{tabular}[c]{@{}l@{}}RDF-\\ ECO\end{tabular} & \begin{tabular}[c]{@{}l@{}}RIDF-\\ ECO\end{tabular} \\ \hline
		AEO     & 0.357    & 0.360                                            & 0.362                                             & 0.383                                             & 0.386                                              & {\color[HTML]{009901} 0.395}                       & {\color[HTML]{00009B} 0.402}                        & {\color[HTML]{FE0000} 0.433}                         \\
		\%Gain  & Baseline & 0.8                                              & 1.4                                               & 7.3                                               & 8.1                                                & {\color[HTML]{009901} 10.6}                        & {\color[HTML]{00009B} 12.6}                         & {\color[HTML]{FE0000} 21.3}                         
		\\ \hline
	\end{tabular}
	
\end{table}


\subsection{Comparison with the State of the Arts}
Here, we demonstrate detailed evaluation results to experimentally validate the efficacy of our contributions in hotistic visual object tracking challenge. We use VOT2016 benchmark during development stage and hyper parameter tuning. To analyze the generalization ability of proposed contributions, we benchmark our trackers with same paramter settings on both VOT2016 and OTB100.
\subsubsection{Evaluation on VOT2016}
We evaluate the top performing models from Table~\ref{eco_comp}, including Ablative trackers of SRDCF, on VOT2016 dataset. As per the results in Table~\ref{vot_comp}, the I-SRDCF, RDF-SRDCF, and RIDF-SRDCF provide a considerable improvement of 3.53\%, 10.60\%, and 11.41\% in AEO, 4.83\%, 17.87\%, and 13.04\% in robustness, respectively. The RIDF-ECO performs favourably against the state-of-the-art trackers including MDNet (won VOT2015) and CCOT (won VOT2016) with a slight improvement of 1.71\% in AEO, and as high as 6.41\% in robustness. The percentage gain is computed relative to baseline.

\begin{table}[!h]
	\centering
	\caption{State-of-the-art comparison on whole VOT2016 dataset.}
	\label{vot_comp}
	\resizebox{\textwidth}{!}{
		\begin{tabular}{lllllllllll}
			\hline
			Trackers                                                            & SRDCF  & \begin{tabular}[c]{@{}l@{}}I-\\ SRDCF\end{tabular} & \begin{tabular}[c]{@{}l@{}}RDF-\\ SRDCF\end{tabular} & \begin{tabular}[c]{@{}l@{}}\textbf{RIDF}-\\ \textbf{SRDCF}\end{tabular} & TCNN &CCOT   & ECO                           & MDNet                         & \begin{tabular}[c]{@{}l@{}}\textbf{RIDF}-\\ \textbf{ECO}\end{tabular} \\  \hline
			AEO                                                                 & 0.1981 & 0.2051                                              & 0.2191                                               & 0.2207                                                & 0.3249 & 0.3310 & {\color[HTML]{009901} 0.3563} & {\color[HTML]{00009B} 0.3584} & {\color[HTML]{FE0000} 0.3624}                       \\
			\begin{tabular}[c]{@{}l@{}}Failure Rate\\ (Robustness)\end{tabular} & 2.07   & 1.97                                                & 1.70                                                 & 1.80                                                  & 0.96 & 0.83   & {\color[HTML]{009901} 0.78}   & {\color[HTML]{00009B} 0.76}   & {\color[HTML]{FE0000} 0.73}                        
			\\ \hline
	\end{tabular}}
	
\end{table}

\subsubsection{Evaluation on OTB100}
As per the evaluation on OTB100 (Table~\ref{otb_comp}), the proposed RIDF-ECO performs favourably against baseline and also the state-of-the-art trackers in most of the existing categories. As per the quantitative study in Table~\ref{otb_comp}, our method finds difficulties in dealing with Background Clutter, Deformation and Out-of-plane rotation. Though
our contributions strengthen the performance of ECO in these failure cases, it is definitely not better than MDNet. Of particular interest, the SRDCF tracker with deep features, i.e. DeepSRDCF lags behind SRDCF with RIDF, i.e. RIDF-SRDCF in success rate as well as precision which are the standard metrics in OTB100 benchmark. Note that exactly same hyper parameters are used in the evaluation on OTB100 as on VOT2016. This verifies the fact that rotation adaptive filters along with other contributions generalizes well across OTB100 and VOT2016 benchmark datasets. 
\begin{table}[!ht]
	\centering
	\caption{State-of-the-art comparison on OTB100 dataset.}
	\label{otb_comp}
	\resizebox{\textwidth}{!}{
		\begin{tabular}{lllllllllll}
			\hline
			Trackers                                                         & \textbf{RIDF-ECO}            & ECO                          & MDNet                        & CCOT                         & \textbf{RIDF-SRDCF} & DeepSRDCF & SRDCF & CFNet & Staple & KCF   \\ \hline
			Out-of-view                                                      & {\color[HTML]{FE0000} 0.767} & {\color[HTML]{3531FF} 0.726} & 0.708                        & {\color[HTML]{009901} 0.725} & 0.712               & 0.619     & 0.555 & 0.423 & 0.518  & 0.550 \\
			Occlusion                                                        & {\color[HTML]{FE0000} 0.721} & {\color[HTML]{3531FF} 0.710} & {\color[HTML]{009901} 0.702} & 0.692                        & 0.652               & 0.625     & 0.641 & 0.573 & 0.610  & 0.535 \\
			\begin{tabular}[c]{@{}l@{}}Illumination\\ Variation\end{tabular} & {\color[HTML]{FE0000} 0.702} & 0.662                        & {\color[HTML]{3531FF} 0.688} & {\color[HTML]{009901} 0.676} & 0.649               & 0.631     & 0.620 & 0.561 & 0.601  & 0.530 \\
			\begin{tabular}[c]{@{}l@{}}Low \\ Resolution\end{tabular}        & {\color[HTML]{FE0000} 0.734} & {\color[HTML]{009901} 0.652} & {\color[HTML]{3531FF} 0.663} & 0.642                        & 0.588               & 0.438     & 0.537 & 0.545 & 0.494  & 0.384 \\
			\begin{tabular}[c]{@{}l@{}}Background\\ Clutter\end{tabular}     & {\color[HTML]{3531FF} 0.648} & {\color[HTML]{009901} 0.638} & {\color[HTML]{FE0000} 0.697} & 0.620                        & 0.634               & 0.616     & 0.612 & 0.592 & 0.580  & 0.557 \\
			Deformation                                                      & {\color[HTML]{3531FF} 0.687} & {\color[HTML]{3531FF} 0.687} & {\color[HTML]{FE0000} 0.722} & {\color[HTML]{009901} 0.657} & 0.652               & 0.645     & 0.641 & 0.618 & 0.690  & 0.608 \\
			\begin{tabular}[c]{@{}l@{}}In-plane\\ rotation\end{tabular}      & {\color[HTML]{FE0000} 0.696} & 0.645                        & {\color[HTML]{3531FF} 0.656} & {\color[HTML]{009901} 0.653} & 0.635               & 0.625     & 0.615 & 0.606 & 0.596  & 0.510 \\
			\begin{tabular}[c]{@{}l@{}}Out-of-plane\\ rotation\end{tabular}  & {\color[HTML]{3531FF} 0.682} & {\color[HTML]{009901} 0.665} & {\color[HTML]{FE0000} 0.707} & 0.663                        & 0.646               & 0.637     & 0.618 & 0.593 & 0.594  & 0.514 \\
			Fast Motion                                                      & {\color[HTML]{FE0000} 0.716} & {\color[HTML]{3531FF} 0.698} & 0.671                        & {\color[HTML]{009901} 0.694} & 0.693               & 0.640     & 0.599 & 0.547 & 0.526  & 0.482 \\ \hline
			\begin{tabular}[c]{@{}l@{}}Overall \\ Success Rate\end{tabular}  & {\color[HTML]{FE0000} 0.702} & {\color[HTML]{3531FF} 0.691} & {\color[HTML]{32CB00} 0.678} & 0.671                        & 0.641               & 0.635     & 0.598 & 0.589 & 0.581  & 0.477 \\
			\begin{tabular}[c]{@{}l@{}}Overall\\ Precision\end{tabular}      & {\color[HTML]{FE0000} 0.937} & {\color[HTML]{3531FF} 0.910} & {\color[HTML]{32CB00} 0.909} & 0.898                        & 0.870               & 0.851     & 0.789 & 0.777 & 0.784  & 0.696 \\ \hline
			
	\end{tabular}}
\end{table}

\subsubsection{Evaluation of CF trackers} \label{ridf-srdcf}
\begin{figure}[!hb]
	\centering
	\includegraphics[height = 5cm,width=0.8\textwidth]{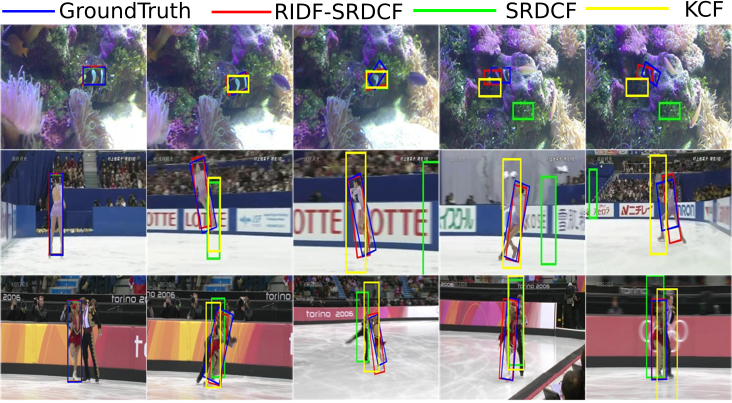}
	\caption{Qualitative analysis of RIDF-SRDCF. The proposed tracker successfully tracks the target under severe rotation, unlike SRDCF and KCF. The rotation adaptive filters assist in determining the orientation of the target object effectively that leads to substantial gain in overall performance. To avoid clumsiness only few bounding boxes are plotted and other variants are quantified in Fig.~\ref{aeo_comp1}.}
	\label{qual_comp1}
\end{figure}

\begin{figure}[!ht]
	\centering
	\includegraphics[height=6cm,width = 0.7\textwidth]{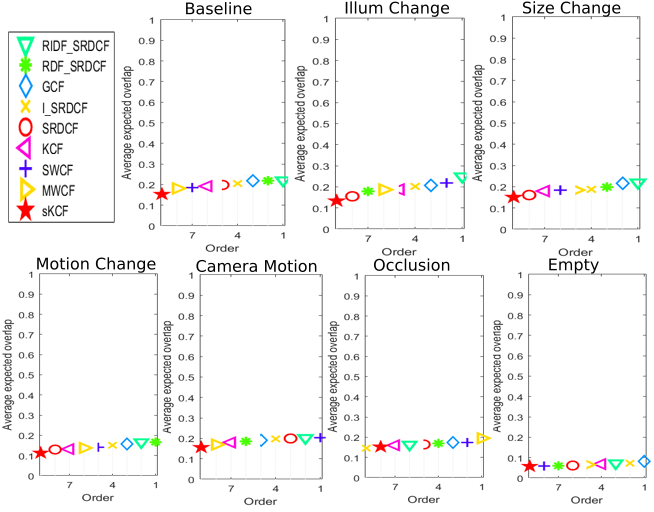}
	\caption{Average Expected Overlap analysis of correlation filter based trackers. }
	\label{aeo_comp1}
\end{figure}

To qualitatively assess the overall performance of RIDF-SRDCF, we compare our results with baseline approach and few other CF trackers on challenging sequences from VOT2016, as shown in Fig. \ref{qual_comp1}. Further, we quantitatively assess the performance by comparing the Average Expected Overlap (AEO) of few correlation filter based trackers, as shown in Fig. \ref{aeo_comp1}.
The proposed RIDF-SRDCF outperforms the standard SRDCF in most of the individual categories that leads to 11.4\% and 13.04\% overall improvement in AEO and robustness, respectively. The categorical comparison, as can be inferred from Fig.~\ref{aeo_comp1}, shows 56.25\%, 23.53\%, 38.46\%, 5.26\%, and 16.66\% gain in Illumination change, Size change, Motion Change, Camera motion, and Empty categories, respectively. Note that the percentage improvement is computed relative to base SRDCF. 
In Table~\ref{rot_comp}, we compare our rotation adaptive scheme with two recent approaches that aims at addressing this issue heuristically. As per our experiments, we report that the proposed rotation adaptive scheme outperforms these counterparts on VOT2016 benchmark. Since base CF trackers are used as core components in most trackers, we believe that the proposed performance gain will be reflected positively in all their derivatives. 

\begin{table}[!h]
	\centering
	\caption{Comparison with two recent rotation adaptive trackers on VOT2016.}
	\label{rot_comp}
	\resizebox{\textwidth}{!}{
		\begin{tabular}{lllll||ll}
			\hline
			Trackers & RAJSSC & SRDCF  & SiameseFC-DSR & \textbf{RIDF-SRDCF}    & ECO & \textbf{RIDF-ECO} \\ \hline
			
			AEO      & 0.1664 & 0.1981 & 0.2084        & 0.2207  &  0.3563     & 0.3624   \\
			\hline
			
	\end{tabular}}
	
\end{table}

\section{Concluding Remarks} \label{cnc}
In this study, we demonstrated that employing a simple, yet effective image enhancement technique prior to feature extraction, can yield considerable gain in tracking paradigm. We analyzed the effectiveness of proposed rotation adaptive correlation filter in standard DCF formulation, and showed compelling results on popular tracking benchmarks. We renovated the sub-grid detection approach by optimizing object's orientation through false positive elimination, which was reflected favourably in the overall performance. Also, the supervision of displacement consistency on CF trackers showed promising results in various scenarios. Moreover, since the DCF formulation is used as backbone of most state-of-the-art trackers, we believe that the proposed rotation adaptive scheme in correlation filters can be suitably integrated into many frameworks and will be useful in boosting the tracking research forward. In future, we will assimilate the performance of proposed tracker on other publicly available datasets~\cite{galoogahi2017need,mueller2016benchmark,smeulders2013visual}.

\bibliographystyle{splncs04}
\bibliography{accv_egbib}

\begin{thebibliography}{10}
\providecommand{\url}[1]{\texttt{#1}}
\providecommand{\urlprefix}{URL }
\providecommand{\doi}[1]{https://doi.org/#1}

\bibitem{bib1}
Adam, A., Rivlin, E., Shimshoni, I.: Robust fragments-based tracking using the
  integral histogram. In: 2006 IEEE Computer Society Conference on Computer
  vision and pattern recognition. vol.~1, pp. 798--805. IEEE (2006)

\bibitem{bib7}
Babenko, B., Yang, M.H., Belongie, S.: Robust object tracking with online
  multiple instance learning. IEEE transactions on pattern analysis and machine
  intelligence  \textbf{33}(8),  1619--1632 (2011)

\bibitem{bib11}
Bolme, D.S., Beveridge, J.R., Draper, B.A., Lui, Y.M.: Visual object tracking
  using adaptive correlation filters. In: Computer Vision and Pattern
  Recognition (CVPR), 2010 IEEE Conference on. pp. 2544--2550. IEEE (2010)

\bibitem{bib2}
Comaniciu, D., Ramesh, V., Meer, P.: Kernel-based object tracking. IEEE
  Transactions on pattern analysis and machine intelligence  \textbf{25}(5),
  564--577 (2003)

\bibitem{bib18}
Danelljan, M., Bhat, G., Khan, F.S., Felsberg, M.: Eco: Efficient convolution
  operators for tracking. In: Proceedings of the 2017 IEEE Conference on
  Computer Vision and Pattern Recognition (CVPR), Honolulu, HI, USA. pp. 21--26
  (2017)

\bibitem{bib15}
Danelljan, M., Hager, G., Shahbaz~Khan, F., Felsberg, M.: Convolutional
  features for correlation filter based visual tracking. In: Proceedings of the
  IEEE International Conference on Computer Vision Workshops. pp. 58--66 (2015)

\bibitem{bib10}
Danelljan, M., Hager, G., Shahbaz~Khan, F., Felsberg, M.: Learning spatially
  regularized correlation filters for visual tracking. In: Proceedings of the
  IEEE International Conference on Computer Vision. pp. 4310--4318 (2015)

\bibitem{bib17}
Danelljan, M., Robinson, A., Khan, F.S., Felsberg, M.: Beyond correlation
  filters: Learning continuous convolution operators for visual tracking. In:
  European Conference on Computer Vision. pp. 472--488. Springer (2016)

\bibitem{bib14}
Danelljan, M., Shahbaz~Khan, F., Felsberg, M., Van~de Weijer, J.: Adaptive
  color attributes for real-time visual tracking. In: IEEE Conference on
  Computer Vision and Pattern Recognition (CVPR), Columbus, Ohio, USA, June
  24-27, 2014. pp. 1090--1097. IEEE Computer Society (2014)

\bibitem{du2015rotation}
Du, Q., Cai, Z.q., Liu, H., Yu, Z.L.: A rotation adaptive correlation filter
  for robust tracking. In: 2015 IEEE International Conference on Digital Signal
  Processing (DSP). pp. 1035--1038. IEEE (2015)

\bibitem{fan2017parallel}
Fan, H., Ling, H.: Parallel tracking and verifying: A framework for real-time
  and high accuracy visual tracking. In: Proc. IEEE Int. Conf. Computer Vision,
  Venice, Italy (2017)

\bibitem{galoogahi2017need}
Galoogahi, H.K., Fagg, A., Huang, C., Ramanan, D., Lucey, S.: Need for speed: A
  benchmark for higher frame rate object tracking. arXiv preprint
  arXiv:1703.05884  (2017)

\bibitem{bib8}
Henriques, J.F., Caseiro, R., Martins, P., Batista, J.: High-speed tracking
  with kernelized correlation filters. IEEE Transactions on Pattern Analysis
  and Machine Intelligence  \textbf{37}(3),  583--596 (2015)

\bibitem{huang2017learning}
Huang, C., Lucey, S., Ramanan, D.: Learning policies for adaptive tracking with
  deep feature cascades. In: IEEE Int. Conf. on Computer Vision (ICCV). pp.
  105--114 (2017)

\bibitem{bib5}
Kalal, Z., Mikolajczyk, K., Matas, J.: Tracking-learning-detection. IEEE
  transactions on pattern analysis and machine intelligence  \textbf{34}(7),
  1409--1422 (2012)

\bibitem{VOT_TPAMI}
Kristan, M., Matas, J., Leonardis, A., Vojir, T., Pflugfelder, R., Fernandez,
  G., Nebehay, G., Porikli, F., \v{C}ehovin, L.: A novel performance evaluation
  methodology for single-target trackers. IEEE Transactions on Pattern Analysis
  and Machine Intelligence  \textbf{38}(11),  2137--2155 (Nov 2016).
  \doi{10.1109/TPAMI.2016.2516982}

\bibitem{bib3}
Kwon, J., Lee, K.M.: Visual tracking decomposition. In: 2010 IEEE Conference on
  Computer Vision and Pattern Recognition (CVPR). pp. 1269--1276. IEEE (2010)

\bibitem{bib9}
Liu, S., Zhang, T., Cao, X., Xu, C.: Structural correlation filter for robust
  visual tracking. In: Proceedings of the IEEE Conference on Computer Vision
  and Pattern Recognition. pp. 4312--4320 (2016)

\bibitem{bib16}
Ma, C., Huang, J.B., Yang, X., Yang, M.H.: Hierarchical convolutional features
  for visual tracking. In: Proceedings of the IEEE International Conference on
  Computer Vision. pp. 3074--3082 (2015)

\bibitem{mueller2016benchmark}
Mueller, M., Smith, N., Ghanem, B.: A benchmark and simulator for uav tracking.
  In: European conference on computer vision. pp. 445--461. Springer (2016)

\bibitem{bib22}
Nam, H., Baek, M., Han, B.: Modeling and propagating cnns in a tree structure
  for visual tracking. arXiv preprint arXiv:1608.07242  (2016)

\bibitem{bib21}
Nam, H., Han, B.: Learning multi-domain convolutional neural networks for
  visual tracking. In: 2016 IEEE Conference on Computer Vision and Pattern
  Recognition (CVPR). pp. 4293--4302. IEEE (2016)

\bibitem{bib12}
Nummiaro, K., Koller-Meier, E., Van~Gool, L.: An adaptive color-based particle
  filter. Image and vision computing  \textbf{21}(1),  99--110 (2003)

\bibitem{bib13}
Oron, S., Bar-Hillel, A., Levi, D., Avidan, S.: Locally orderless tracking.
  International Journal of Computer Vision  \textbf{111}(2),  213--228 (2015)

\bibitem{bib19}
Petrou, M., Petrou, C.: Ch.4: Image enhancement. Image Processing: The
  Fundamentals pp. 293--394 (2010)

\bibitem{rout2018rotation}
Rout, L., Sidhartha, Manyam, G.R., Mishra, D.: Rotation adaptive visual object
  tracking with motion consistency. In: 2018 IEEE Winter Conference on
  Applications of Computer Vision (WACV). pp. 1047--1055 (March 2018)

\bibitem{smeulders2013visual}
Smeulders, A.W., Chu, D.M., Cucchiara, R., Calderara, S., Dehghan, A., Shah,
  M.: Visual tracking: An experimental survey. IEEE Transactions on Pattern
  Analysis \& Machine Intelligence (1) (2013)

\bibitem{bib6}
Wang, S., Lu, H., Yang, F., Yang, M.H.: Superpixel tracking. In: 2011 IEEE
  International Conference on Computer Vision (ICCV). pp. 1323--1330. IEEE
  (2011)

\bibitem{WuLimYang13}
Wu, Y., Lim, J., Yang, M.H.: Online object tracking: A benchmark. In: IEEE
  Conference on Computer Vision and Pattern Recognition (CVPR) (2013)

\bibitem{zhang2015joint}
Zhang, M., Xing, J., Gao, J., Shi, X., Wang, Q., Hu, W.: Joint scale-spatial
  correlation tracking with adaptive rotation estimation. In: Proceedings of
  the IEEE International Conference on Computer Vision Workshops. pp. 32--40
  (2015)

\bibitem{bib4}
Zhang, T., Liu, S., Ahuja, N., Yang, M.H., Ghanem, B.: Robust visual tracking
  via consistent low-rank sparse learning. International Journal of Computer
  Vision  \textbf{111}(2),  171--190 (2015)

\end{thebibliography}

\end{document}